\def\year{2021}\relax
\documentclass[letterpaper]{article} 
\usepackage{aaai21}  
\usepackage{times}  
\usepackage{helvet} 
\usepackage{courier}  
\usepackage[hyphens]{url}  
\usepackage{graphicx} 
\usepackage{natbib}  
\usepackage{caption} 
\usepackage[ruled,vlined,linesnumbered]{algorithm2e}
\usepackage{amssymb}
\usepackage{subfigure}
\usepackage{multirow}

\title{Euclidean Distance-Optimal Post-Processing of Grid-Based Paths}

\author{
    Guru Koushik Senthil Kumar\textsuperscript{\rm 1}, Sandip Aine\textsuperscript{\rm 2}, Maxim Likhachev\textsuperscript{\rm 2}
    \\
}
\affiliations{
    \textsuperscript{\rm 1}Department of Mechanical Engineering, Carnegie Mellon University\\
    \textsuperscript{\rm 2}Robotics Institute, Carnegie Mellon University\\

    \{gsenthil, asandip\}@andrew.cmu.edu, maxim@cs.cmu.edu

}

\begin{document}

\maketitle

\begin{abstract}
Paths planned over grids can often be suboptimal in an Euclidean space and contain a large number of unnecessary turns. Consequently, researchers have looked into post-processing techniques to improve the paths after they are planned. In this paper, we propose a novel post-processing technique, called Homotopic Visibility Graph Planning (HVG) which differentiates itself from existing post-processing methods in that it is guaranteed to shorten the path such that it is at least as short as the provably shortest path that lies within the same topological class as the initially computed path. We propose the algorithm, provide proofs and compare it experimentally against other post-processing methods and any-angle planning algorithms.
\end{abstract}

\section{Introduction}
\noindent Computing shortest paths in a continuous 2D environment has been of interest for researchers in various domains such as robotics, game development and computational geometry. \cite{vgraph} solved the problem of finding the shortest path among polygonal obstacles using visibility graphs. In visibility graphs, search is performed over a graph with vertices at convex obstacle corners and guarantees to return Euclidean optimal paths in two dimensional spaces with polygonal obstacles. However, constructing and planning on visibility graphs can become slow as the number of obstacles increases. In computational geometry literature, \cite{polygoncutting,besp2004} solved the Euclidean shortest path problem with the funnel algorithm but the algorithm requires triangulation of the environment which is a not an efficient way to represent environments that can vary dynamically. Consequently, grid representation of the environment dominated search-based planning literature due to its ease of use and flexibility in representation for varying costs and dynamically changing environments. The limitations of using grid-based graphs for search-based planning algorithms such as A* and its derivatives is that the angle of traversal is limited by increments of 45$^\circ$ (assuming an eight-connected grid). As a result, these planners can produce unrealistic looking paths with unnecessary turns and high path cost. The need to alleviate the suboptimal costs and unnecessary turns of the generated paths has lead to two research directions: post-processing of computed paths and any-angle planning algorithms. 

In this work, we present a novel algorithm for post-processing grid-based paths which returns a path that is at least as short as the Euclidean-distance optimal path in the homotopy class of the original path. We utilize the fact that search over visibility graphs gives Euclidean optimal paths by building a `local' Homotopic Visibility Graph around the grid-search path and performing a search over it to obtain the post-processed path. Considering that visibility graph search can become time consuming due to high branching factor, there is a need to prune the maximum number of vertices and edges possible. Our algorithm chooses only convex obstacle corners in the homotopy class of the grid path that could potentially contribute to a taut path, to build the visibility graph. The process of finding the relevant obstacle corners and building the visibility graph is highly parallelizable and hence has a much better runtime compared to planning on full visibility graphs.

In the following sections of the paper the related work is detailed, the Homotopic Visibility Graph Planning (HVG) algorithm is detailed and proofs showing that the post-processed path is at least as short as the provably optimal path lying within the same homotopy class as the initial grid-based path are provided. The results of experiments on grid-based maps along with comparison to other post processing algorithms and any-angle planning algorithms are presented. HVG is able to scale significantly better than any-angle algorithms to large and dense maps and achieves better runtimes while providing a homotopic optimality guarantee.

\section{Related Work}

A two step approach of path computation and post-processing of grid-based paths has been commonly used to reduce the cost of the generated paths. The most commonly used technique is greedy post-processing \cite{highqualitypaths} where three consecutive nodes are taken and if there is a line of sight between the first and the third node, the second node is removed from the path. This is followed until there are no path shortcuts available. The main pitfall of such greedy post-processing is that it only removes nodes from the path and does not allow for addition of nodes which is necessary for obtaining a provably shortest path. \cite{stringpull} introduced a post-processing algorithm which allowed for addition and removal of nodes in the path and always generated post-processed paths with no heading changes in freespace. However, to the best of our knowledge there still exists no post-processing algorithm that provides guarantees on optimality or bounded sub-optimality of the post-processed path in grid-based representations.

Any-angle algorithms \cite{uras2015empirical} such as Theta* \cite{thetastar}, ANYA \cite{anya} and Field D* \cite{fielddstar} resolve the problem of angle limitations in search over grid-based graphs by interleaving shortcutting with search and propagating information along grid edges without constraining the path to go along the grid edges. Among any-angle algorithms, the approaches followed to alleviate unnecessary turns differs from one algorithm to another. The Theta* algorithm interleaves the path shortcutting step during expansion of nodes by checking for a line of sight between the node and its parent's predecessor. Whereas ANYA performs the search over row-wise intervals on the grid-based graph to restrict the intermediate nodes of the path to obstacle corners. Although any-angle planning algorithms give shorter paths than A*, they are typically slower and do not provide optimality guarantees with the exception of ANYA \cite{anya}.

The problem of post-processing a coarse path to obtain Euclidean optimal paths in the same homotopy class is well studied in computational geometry literature. \cite{polygoncutting, besp2004} introduced the funnel algorithm which gives the shortest path in the homotopy class with respect to the Euclidean or link metric. The pitfall of this approach is the requirement of the boundary-boundary triangulation of the map which is a non-trivial endeavor. 

The motivation for this work comes from the need to have a post-processing algorithm with provable guarantees in the grid-based representation of 2D environments as it is widely used in multiple domains. Existing post-processing algorithms do not provide theoretical bounds and any-angle algorithms do not scale well to large maps and maps with high obstacle density. The proposed Homotopic Visibility Graph (HVG) algorithm addresses the mentioned gaps in research by providing a Euclidean optimality guarantee in the homotopy class and scales well to large and dense maps due to its ability to be parallelized.

\section{Terminology}
\noindent A grid representation of a 2D environment is a discretization of the space into equally sized cells which are either obstacles or not. A graph $G = [S,E]$ is constructed with corners of the cells as the graph vertices and the edges of the cells as the edges of the graph $E$ (either 4-connected or 8-connected). Let $S$ be the set of all vertices in the graph and $s,g \in S$ where $s=[s_x,s_y]$ and $g=[g_x,g_y]$ denote the start and goal vertex of the path. The set of convex obstacle corners in the grid representation of a planar map is denoted by $V \subseteq S$. From Lozano-P\'{e}rez and Wesley (1979), if a vertex $v \in P_{opt}$ where $P_{opt}$ is the Euclidean optimal path, then $v \in \{s, g\} \cup V$.

\noindent \textbf{Grid-based Path: }A grid-based path on graph $G$ is defined as $P=\{s=p_1, p_2, ... p_n=g\}$ if $s \in S$ is the start vertex and $g \in S$ is the goal vertex and $p_i \in S$ with consecutive vertices in the path connected by edge $e_i \in E$.

\noindent \textbf{Line of Sight (LOS): } The graph vertices $p,q \in S$ are said to have line of sight if the line joining the vertices in the Cartesian space does not intersect any obstacle.

\noindent \textbf{Visibility Graph: }A graph $VG = [V, E_{vg}]$ is defined as a visibility graph if $V \subseteq S$ is the set of all convex obstacle corners in $S$ and $E_{vg}$ is the set of the lines joining all pairs of vertices $v_i, v_j \in V$ which have line of sight.

\noindent \textbf{Homotopic Trajectories: } Formally, homotopic trajectories are defined as two trajectories $\tau_1$ and $\tau_2$ with the same start and end coordinates $s$ and $g$ iff one can be continuously deformed into the other without intersecting an obstacle.
\[ \tau_1, \tau_2 : [0,1] \rightarrow \mathbb{R}^2 \]
\[\tau_1(0) = \tau_2(0) = s, \tau_1(1) = \tau_2(1) = g\]
then, $\tau_1$ and $\tau_2$ are homotopic iff there exists a continuous map $\eta$ such that
\[\eta : [0,1]\times[0,1] \rightarrow \mathbb{R}^2\]
\[\eta(\alpha,0) = \tau_1(\alpha) \forall \alpha \in [0,1]\]
\[\eta(\beta,1) = \tau_2(\beta) \forall \beta \in [0,1]\]
\[\eta(0,\gamma) = s, \eta(1,\gamma) = g \forall \gamma \in [0,1]\]

\begin{figure}[h!]
    \centering
    \includegraphics[width=\linewidth]{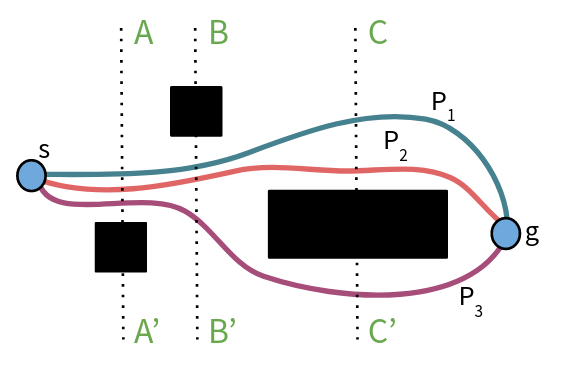}
    \caption{The paths P$_1$ and P$_2$ and homotopic trajectories with canonical sequence AB'C whereas P$_3$ lies in a different homotopy class with canonical sequence AB'C'}
    \label{fig:homotopy}
\end{figure}

\noindent \textbf{A test for homotopic paths: }As shown in \cite{testhomotopy}, two paths in a plane belong to the same homotopy class if and only if they have the same reduced canonical sequence. To determine the canonical sequence of a path, we begin by choosing one representative point for each obstacle in the map and draw an infinite vertical line passing through each of those points. Each of these lines are treated as two separate rays originating from the obstacle's representative point and each ray is given a unique ID. The canonical sequence is then constructed by appending the ID's of the rays which are intersected by the path in order. The canonical sequence is further reduced to eliminate redundant crossings. Two paths are then said to be homotopic iff they have the same reduced canonical sequence (see Figure \ref{fig:homotopy}).

\noindent \textbf{Euclidean Optimal Path: }Performing an optimal search on the graph $G$ will lead to a path that is optimal with respect to the grid resolution. A path from $s \in S$ to $g \in S$ is said to be a Euclidean optimal path if it minimizes the Euclidean distance metric $d(p,q) = ||p-q||_2$.

\begin{figure}[h!]
    \centering
    \includegraphics[width=\linewidth]{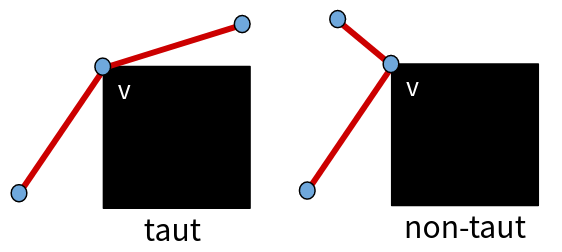}
    \caption{The figure illustrates taut and non-taut exits at an obstacle corner. The taut exit subtends an angle less than 180$^\circ$ whereas the non-taut exit subtends an angle greater than 180$^\circ$. The taut path cannot be shortened without intersecting an obstacle. The non-taut path can be shortened by removing the intermediate vertex $v$.}
    \label{fig:taut}
\end{figure}

\noindent \textbf{Taut Path: }A 2D path on a discrete grid is a taut path if it has no turns in freespace and all of its intermediate vertices are convex obstacle corners. Additionally, all the pairs of path segments intersecting at an obstacle corner should subtend an angle less than $180^{\circ}$ at the obstacle corner. Euclidean optimal paths are always taut paths with vertices at obstacle corners as shown in the work on visibility graphs \cite{vgraph}. A taut exit (see Figure \ref{fig:taut}) is one which subtends an angle less than 180$^\circ$ at an obstacle corner. A taut path is one where each intermediate vertex is a taut exit as illustrated in Figure \ref{fig:taut}.

\section{HVG Algorithm}

\begin{figure*}[h!]
  \centering
  \subfigure[The initially computed grid-based path]{\includegraphics[width=0.48\textwidth]{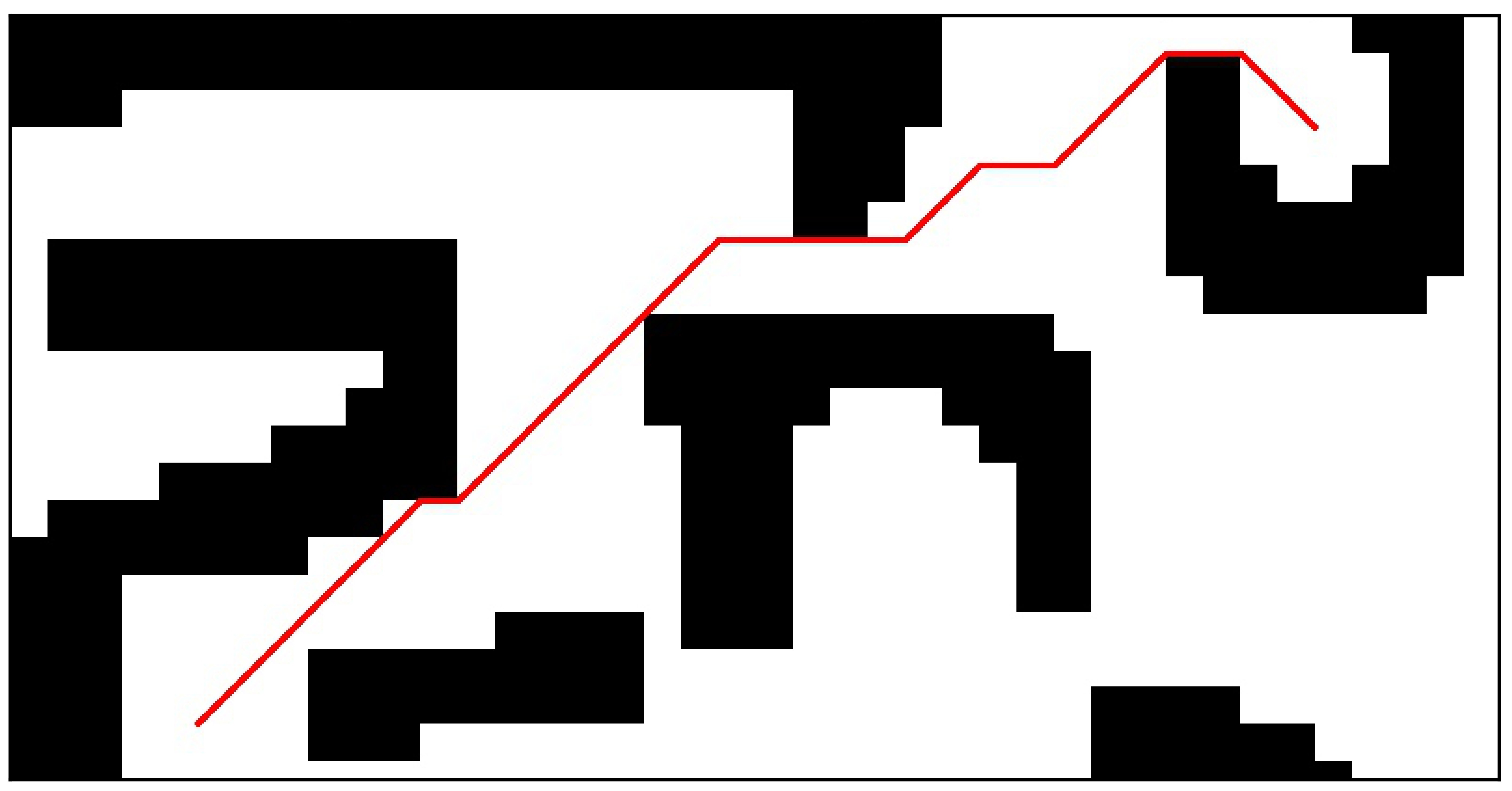}}\quad
  \subfigure[The blue circles indicate the vertices that were encountered when performing scans in the four directions]{\includegraphics[width=0.48\textwidth]{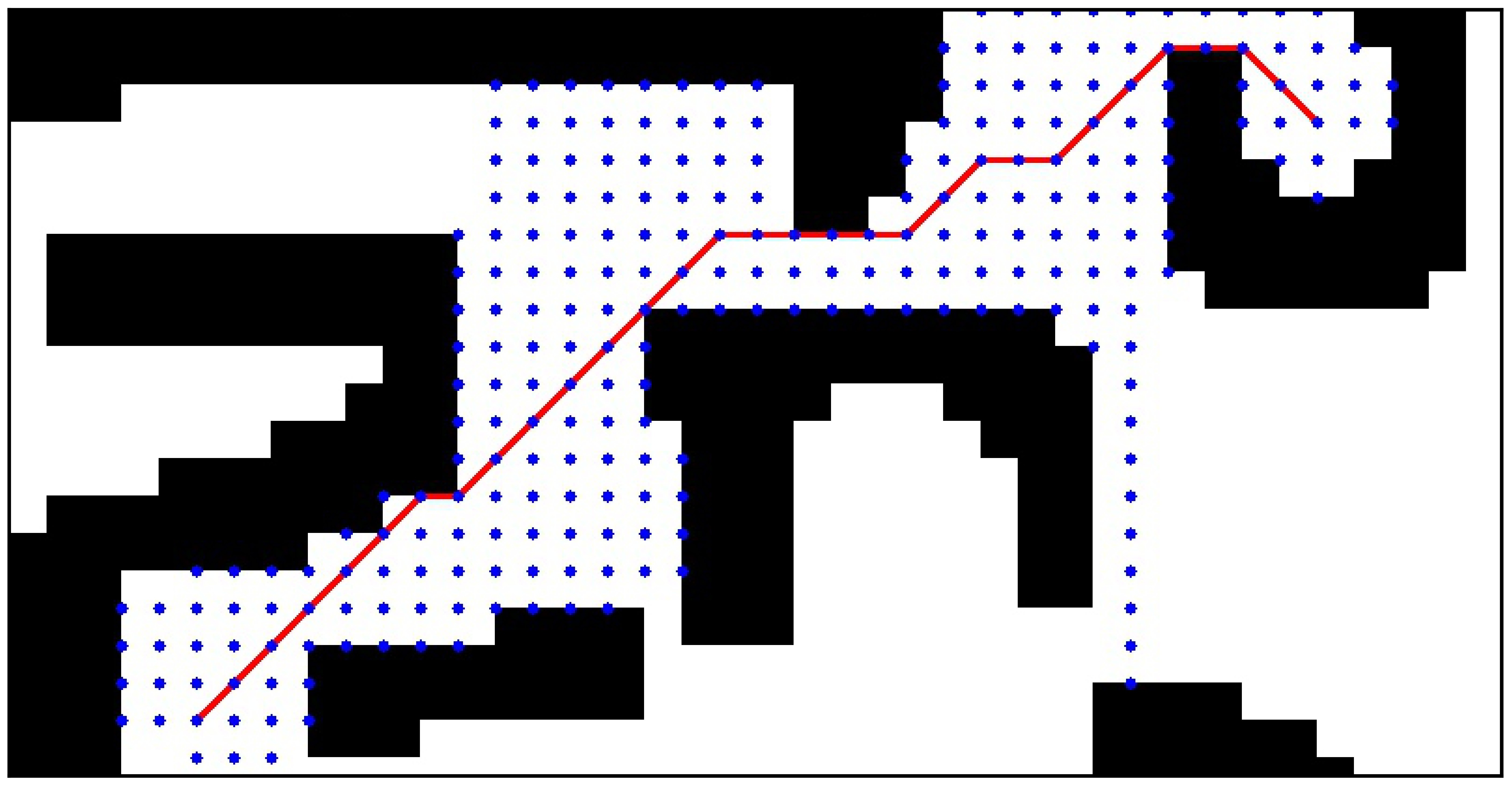}}
  \subfigure[The orange circles indicate the vertices of the Homotopic Visibility Graph $HVG$ computed after completing the scanning]{\includegraphics[width=0.48\textwidth]{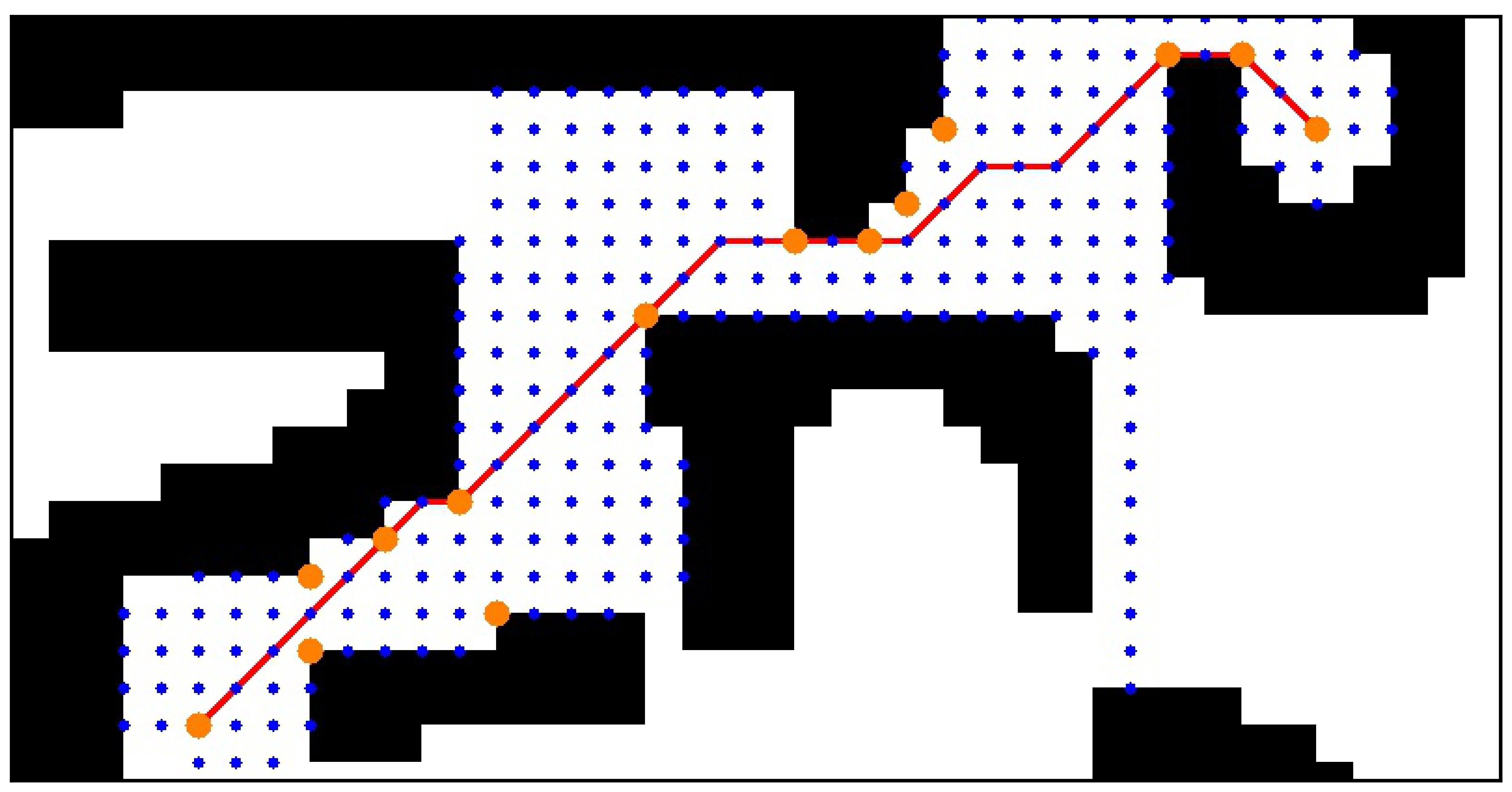}}\quad
  \subfigure[The green path is the post-processed path $\hat{P}$ and the blue lines illustrate the constructed Homotopic Visibility Graph $HVG$]{\includegraphics[width=0.48\textwidth]{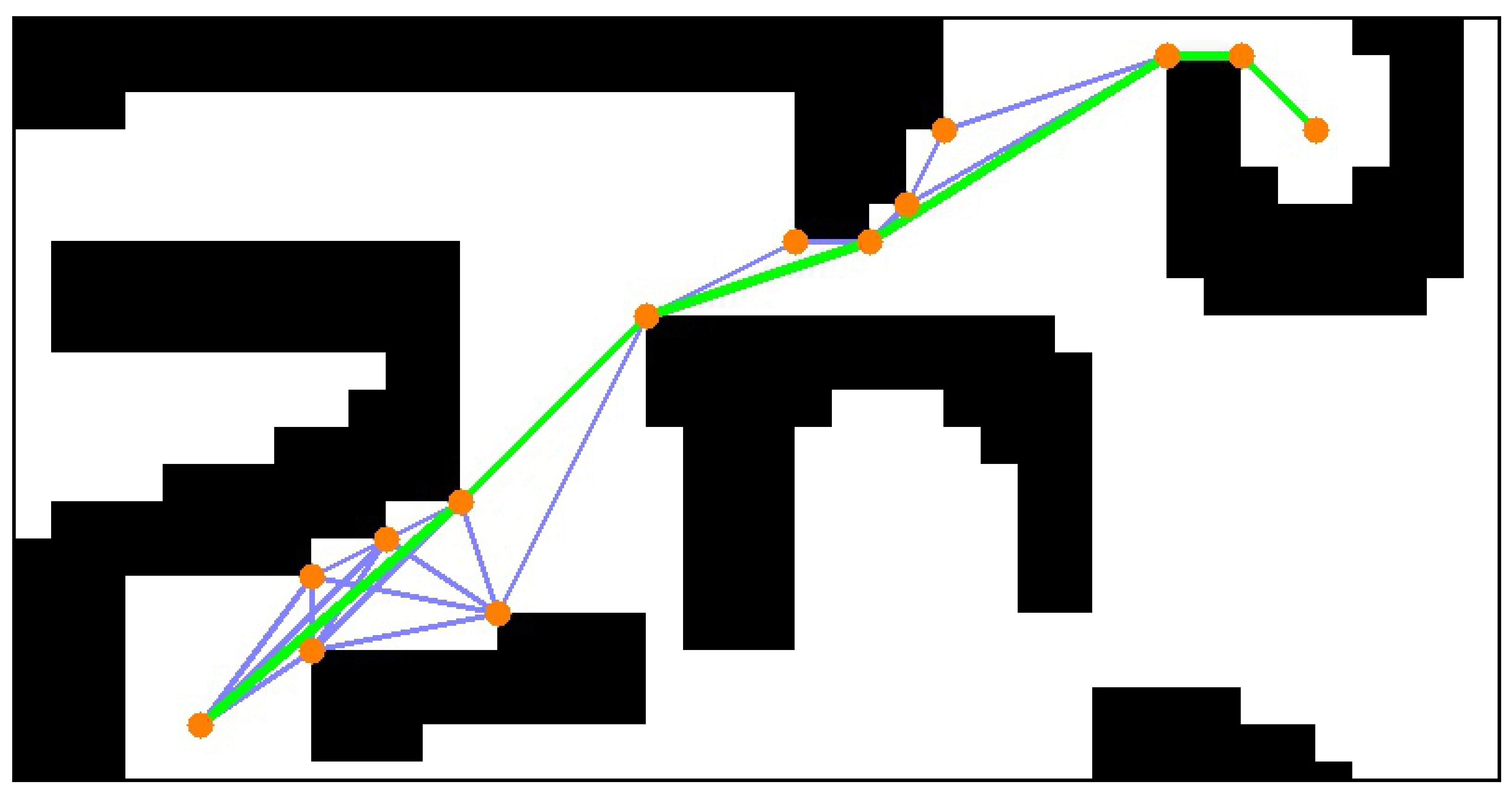}}
  \caption{Illustration of the HVG post-processing algorithm where a) shows the initial grid-based path which has unnecessary turns in freespace. b) shows the vertices encountered when performing scans originating from each vertex of the grid-based path in the up, down, left and right directions. c) illustrates the vertices that are common in the $V_v$ and $V_h$ sets as described in step 24 of Algorithm 1. d) shows the constructed visibility graph and the post-processed path that is Euclidean-distance optimal in its homotopy class.}
  \label{fig:vis}
\end{figure*}

The Homotopic Visibility Graph Planning (HVG) algorithm proposed in this paper is a post-processing algorithm that takes as input the grid-based path $P = (s=p_1,p_2,p_3 ... p_n=g)$ and returns the post-processed path $\hat{P}$. The core idea behind the approach utilizes two properties of Euclidean distance-optimal paths in 2D environments: visibility graphs provide Euclidean shortest paths and the shortest paths are always taut. If a complete visibility graph $VG$ can be constructed within the homotopy class of $P$, then the shortest path can be obtained by performing an optimal search in $VG$. However $VG$ can be pruned to reduce the number of edges and vertices by exploiting the knowledge of the grid-based path $P$. In our algorithm, we construct $HVG \subseteq VG$ from $P$ as described in Algorithm 1. A visualization of the algorithm is shown in Figure \ref{fig:vis}.

\begin{algorithm}[h]
\SetAlgoLined
\SetKwInOut{Input}{Input}
\SetKwInOut{Output}{Output}
\Input{Grid path $P = (s=p_1,p_2, ... p_n=g)$}
\Output{Post-processed path $\hat{P}$}
$V_h = \emptyset$, $V_v = \emptyset$, $V_{HVG} = \{s,g\}$\\
\For{node in P}{
\If{iscorner(node)}{
$V_{HVG}$.append(node)\;
}
\For{dir in \{up,down,left,right\}}{
obstacle\_hit = false\;
curr = node\;
\While{obstacle\_hit == false}
{
    new = curr + dir\;
    obstacle\_hit = CollisionDetect(curr, new)\;
    curr = new\;
    \If{iscorner(curr)}{
    \uIf{dir == left or dir == right}{
    $V_{h}$.append(curr)\;
    }
    \Else
    {$V_{v}$.append(curr)\;}
    break\;
    }   
}
}
}
$V_{HVG} = V_{HVG} \cup (V_h \cap V_v)$\;
$HVG$ = VisibilityGraph($V_{HVG}$)\;
$\hat{P}$ = Search($V_{HVG}$, $p_1$, $p_n$)\;
\textbf{return} $\hat{P}$
\caption{Homotopic Visibility Graphs (HVG)}
\end{algorithm}

\noindent \textbf{Finding vertices of $HVG$: }
A key observation that aids in finding the vertices of $HVG$ is that $v \in V$ is a candidate vertex for the optimal path $\hat{P}$ in the homotopy class of $P$ \textit{if and only if} it provides a taut exit (see Figure \ref{fig:taut}). For every node $p \in P$, a line of sight scan originating from $p$ is performed in all the four directions parallel to the axes of the grid. Two lists $V_h, V_v$ are maintained for storing obstacle corner vertices encountered during horizontal scans and vertical scans. The scans are performed until an obstacle or an obstacle corner is encountered. In the case that an obstacle is encountered, the scan in that direction is terminated. If an obstacle corner is encountered, then we add that to the corresponding list and the scan is terminated. After performing scans for all the nodes $p \in P$, the $HVG$ vertices are determined by $V_{HVG} = V_h \cap V_v$. By finding obstacle corner vertices that have a horizontal and vertical line of sight scan from the grid-search path, we ensure that only vertices that can potentially be a part of a taut exit are included. Since the scans originating at node $p_i \in P$ does not depend on scans from any other node $p_j \in P$, the scans can be parallelized. The proof for optimality is provided in the section below.

\noindent \textbf{Post-processed path $\hat{P}$: }The post-processed path $\hat{P}$ is simply obtained by performing an optimal search over the constructed visibility graph $HVG$. The resulting path may lie in neighboring homotopy classes since there is no restriction for the path to lie in the same homotopy class as $P$. However, the post-processed path $\hat{P}$ is guaranteed to be at least as short as the Euclidean optimal path in the homotopy class of $P$. The construction of the visibility graph $HVG$ requires pairwise line of sight scans which can be parallelized for efficient computation. 
\begin{table}[h!]
\centering
\begin{tabular}{|l|l|l|}
\hline
\multicolumn{3}{|c|}{Grid Path: E1,D2,D3,C4,B5,B6,B7,B8} \\ \hline
\multicolumn{1}{|c|}{\textbf{Grid Path Node}} & \multicolumn{1}{c|}{\boldmath{$V_h$}} & \multicolumn{1}{c|}{\boldmath{$V_v$}} \\ \hline
E1 & E5 & C1 \\ \hline
D2 & E5 & C1 \\ \hline
D3 & E5 & C1,C3 \\ \hline
C4 & E5,C3,C5 & C1,C3 \\ \hline
B5 & E5,C3,C5 & C1,C3,C5 \\ \hline
B6 & E5,C3,C5 & C1,C3,C5 \\ \hline
B7 & E5,C3,C5 & C1,C3,C5 \\ \hline
B8 & E5,C3,C5 & C1,C3,C5 \\ \hline
\multicolumn{3}{|c|}{$V_{HVG}$ = {[}E1,C3,C5,B8{]}, HVG Path = E1,C5,B8} \\ \hline
\end{tabular}
\caption{An execution of HVG algorithm. For each node in the grid based path, the state of the sets $V_h$ and $V_v$ is shown.}
\label{table:exhvg}
\end{table}

\begin{figure}[h!]
    \centering
    \includegraphics[width=0.9\linewidth]{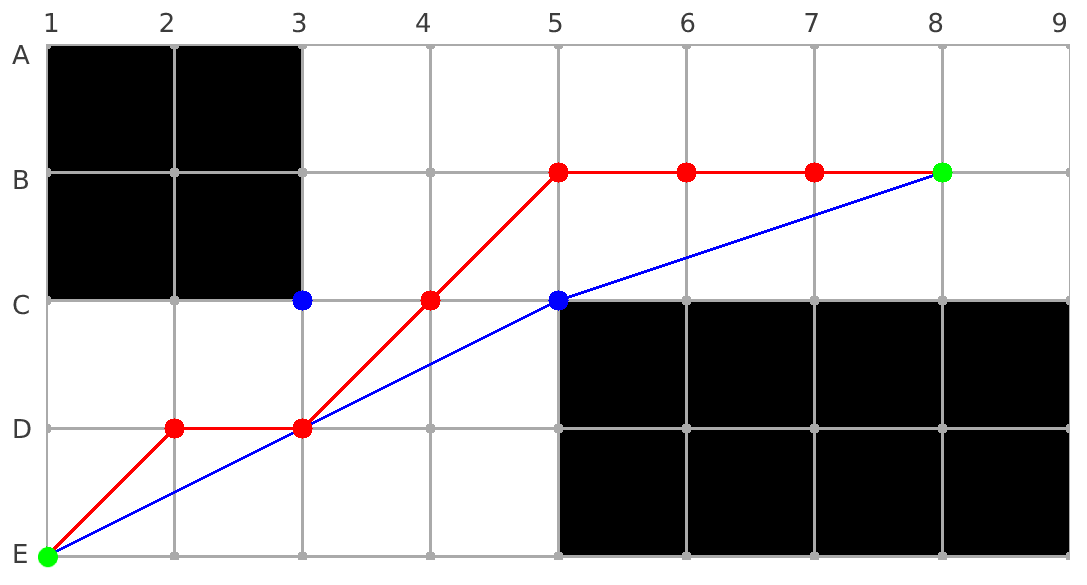}
    \caption{Illustration of the execution of HVG algorithm. The green circles represent the start and goal vertices. Red circles denote the vertices of the grid-based path $P$ which is computed using A*. Blue path is the post-processed path $\hat{P}$ with the blue circles denoting the HVG vertices $V_{HVG}$.}
    \label{fig:example}
\end{figure}

\noindent \textbf{Execution of HVG: }An example execution of HVG is illustrated in Figure \ref{fig:example} and Table \ref{table:exhvg}. The green circles denote the start and goal vertices. The red path is the initially computed grid-based path $P$ where the red circles represent the vertices of the path. The blue path is the post-processed path $\hat{P}$ with the blue circles denoting the vertices $V_{HVG}$ found by HVG. For each node in $P$, Table \ref{table:exhvg} shows the state of the lists $V_v$ and $V_h$. The horizontal and vertical scans are performed from the vertices colored in red or green and the result is agnostic to the order in which the scans are performed.

\section{Proof}
\noindent In this section, the proof for homotopic optimality of the post-processing algorithm is presented. The outline of the proof is as follows. For the given grid-based path $P$ in homotopy class $H$, let $P_{opt} \in H$ be the optimal path in $H$ and $V_{HVG}$ be the set of vertices found by the HVG algorithm. If we can show that $P_{opt} \subseteq V_{HVG}$ then the visibility graph constructed using $V_{HVG}$ will be guaranteed to contain the path $P_{opt}$. The aim is to show that an obstacle corner vertex $v^i \in V$ belongs to the optimal path in the homotopy class $P_{opt}$ if and only if there exists a grid search node $p_h^i, p_v^i \in P$ such that $p_h^i$ has a horizontal scan to $v$ and $p_v^i$ has vertical scan to $v$. Here, a scan is defined as a line that passes through freespace and does not go along obstacle edges or intersect an obstacle.



\noindent \textbf{Lemma 1: }For the grid-based path $P \in H$, consider the scenario when the Euclidean-distance optimal path $P_{opt} \in H$ has only one intermediate vertex $v$. For $v \in V$ to be a taut exit belonging to the optimal path, the angle subtended by the path segments at $v$ is less than $180^{\circ}$ and the vertex $v = [v_x,v_y]$ satisfies 

\[s_{x} \leq v_x \leq g_{x}\] 
\[s_{y} \leq v_y \leq g_{y}\]


\noindent \textbf{Proof: }Let us assume that $v_y < s_{y} < g_{y}$ and $v$ is part of the optimal path. But this would lead to the angle subtended at $v$ by the path segments to be greater than $180^{\circ}$ which would make it a non-taut exit. This contradicts the assumption that $v$ is part of the optimal path. Similarly, violation of the other inequalities can be shown to lead to $v$ not being part of the optimal path. Hence, it is proven by contradiction that for a Euclidean optimal path with two segments, containing an obstacle corner $v$ as one of the vertices, the inequalities $s_{x} < v_x < g_{x}$ and $s_{y} < v_y < g_{y}$ are satisfied.

\noindent \textbf{Lemma 2: }For any grid-based path $P$ that is homotopic to $P_{opt}$ with one intermediate vertex $v \in P_{opt}$, there exists $p_h, p_v \in P$ such that 
\[p_{hy} = v_y\] 
\[p_{vx} = v_x\] 

\noindent and the line joining the pair of points $\{v,p_h\}$, $\{v,p_v\}$ are collision free.


\begin{figure}[h!]
    \centering
    \includegraphics[width=\linewidth]{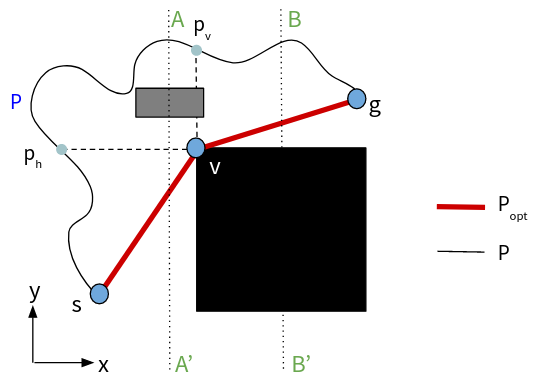}
    \caption{Illustration to show that the homotopy class changes if an obstacle is present between either pair of points $\{v,p_h\}$ and $\{v,p_v\}$}
    \label{fig:homotopyproof}
\end{figure}

\noindent \textbf{Proof: }The grid search path $P$ is homotopic to the path $P_{opt}$. In Lemma 1, we stated and proved that if $v \in P_{opt}$, then $s_{x} < v_x < g_{x}$ and $s_{y} < v_y < g_{y}$. Since the grid search path $P$ is continuous between $s$ and $g$, there exists $p_h, p_v \in P$ such that $p_{hy} = v_y$ and $p_{vx} = v_x$. Let us assume that the either of the pair of points $\{v,p_h\}$ and $\{v,p_v\}$ are obstructed by an obstacle. As shown in Figure \ref{fig:homotopyproof}, the canonical sequence \cite{testhomotopy} for testing homotopy for $P$ is $AB$ and for $P_{opt}$ is $A'B$ which indicates that the grid search path $P$ and $P_{opt}$ belongs to different homotopies. Thus, by contradiction we prove that the pair of points $\{v,p_h\}$ and $\{v,p_v\}$ has to be collision free.

\noindent \textbf{Lemma 3: }When $P_{opt}$ is the optimal path in $H$, there exists $p_h^i,p_v^i \in P$ for every $v^i \in P_{opt}/\{s,g\}$ such that $p_{hy}^i = v_y^i$ and $p_{vx}^i = v_x^i$ and the line joining the pair of points $\{v^i,p_h^i\}$ and $\{v^i,p_v^i\}$ are collision free.

\noindent \textbf{Proof: }We generalize Lemma 1 and Lemma 2 for $P_{opt}$ with multiple intermediate vertices. The two step proof for the lemma is as follows:
\begin{enumerate}
    
    \item First we show that when the paths $P$ and $P_{opt}$ intersect at only two points namely the start $s$ and goal $g$ coordinates, there exists horizontal and vertical line of sight scans from $v_i \in P_{opt}$ to $P$. We utilize the property that any subset of the optimal path is also optimal. Consider the sub-path $P' = \{v_{i-1}, v_i, v_{i+1}\}$ from $P_{opt} = (s=v_1,v_2, ... v_k=g)$ where $P'$ is an optimal path with start $v_{i-1}$ and goal $v_{i-1}$ with only one taut exit as shown in Lemma 2. Construct another path $P'' \in H$ such that $P'' = \{v_{i-1}, v_{i-2} .. v_1, P, v_k, v_{k-1} ... v_{i+1}\}$. From Lemma 2, $P'$ and $P''$ are homotopic trajectories and the existence of $p_h^i,p_v^i \in P''$ can be shown. Further, it holds that $p_h^i,p_v^i \in P$ because if $p_h^i,p_v^i \in P''/P$ then that would be a valid path shortcut in $P_{opt}$ which contradicts the assumption that $P_{opt}$ is the optimal path in $H$. Hence proven that for every $v_i \in P_{opt} \: \exists \: p_h^i,p_v^i \in P$
    
    \item In the case that the paths $P$ and $P_{opt}$ intersect at more than two points, the paths are split at the intersection points and for each of the segments, (1.) of the proof holds.
\end{enumerate}

\noindent \textbf{Theorem 1: }For a given grid-based path $P$ in homotopy class $H$ with $P_{opt}$ being the Euclidean-distance optimal path in $H$, the post-processed path $\hat{P}$ returned by the Homotopic Visibility Graph (HVG) algorithm is such that $length(\hat{P}) \leq length(P_{opt})$.

\noindent \textbf{Proof: }Vertices of the graph found in lines 2-24 of Algorithm 1 is a superset of the vertices $v \in P_{opt}$ as stated and proved in Lemma 3. Hence, the path found by performing an optimal search on the visibility graph constructed using the found vertices is guaranteed to be at least as short as $P_{opt}$.

\section{Experiments}

For benchmarking the algorithms, we use city and random maps from \cite{movingai} ranging from sizes 512x512 to 6000x6000. To test the algorithms on maps with high obstacle density, custom maps were generated by randomly sampling obstacles. We test the path length and runtime for A$^*$ with HVG, A$^*$ with Greedy post-processing (G-PP), A$^*$ with String Pulling (SP), Theta$^*$ and ANYA. All the algorithms were implemented in C++ on an Intel i7-6700 CPU (3.40GHz) and 32GB RAM. The implementations for Theta$^*$ \& ANYA is from \cite{uras2015empirical}. 

\begin{figure}[h!]
  \centering
  \includegraphics[width=\linewidth]{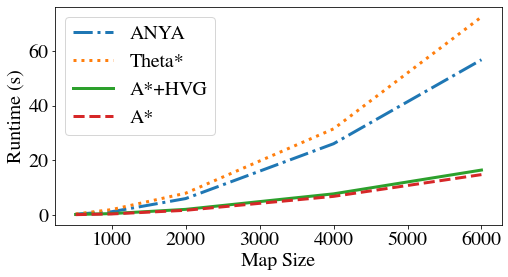}
  \caption{Runtime vs Map Size statistics for maps generated with 40\% obstacle density}
  \label{fig:mapsize}
\end{figure}

\noindent \textbf{Runtime vs Map Size: }The scalability of HVG and other algorithms are tested in maps of size 512x512 to 6000x6000. The practical merits of using A*+HVG in place of ANYA or Theta* is seen in Figure \ref{fig:mapsize} where the scalability of the proposed algorithm (A*+HVG) to large maps is shown. On average, A*+HVG is four times faster than Theta* and three times faster than ANYA in maps of size 6000x6000. While the time taken for HVG is comparable to that of the search itself in maps of size 512x512, the ratio of time taken by HVG to that of A* grows significantly smaller as the map size increases. 

\begin{figure}[h!]
  \centering
  \includegraphics[width=\linewidth]{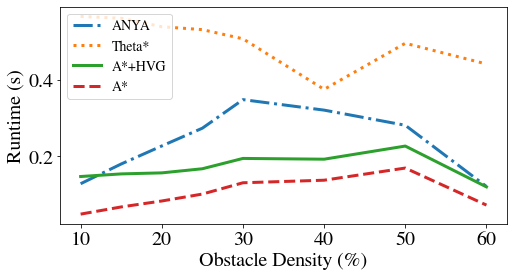}
  \caption{Runtime vs Obstacle Density statistics for maps of size 512x512 generated with randomly sampled obstacles}
  \label{fig:obsdensity}
\end{figure}

\noindent \textbf{Runtime vs Obstacle Density: }The performance of A*+HVG is benchmarked in maps of varying obstacle densities and the runtime is plotted in Figure \ref{fig:obsdensity}. It can be seen that A*+HVG performs better than the other algorithms in maps of high obstacle densities. ANYA exploits freespace regions efficiently by searching over the space of intervals instead of neighboring vertices in the grid-based graph and hence performs better in maps which are less dense.

\noindent \textbf{Path Cost: }The path quality generated by HVG when coupled with A* is comparable to that of Theta* and ANYA. The length of paths generated by A*+HVG is shorter than that of A*, A*+G-PP and A*+SP. Theta* and ANYA generate marginally shorter paths than A*+HVG which is a consequence of the A* path being in a homotopy class that does not contain the globally Euclidean-distance optimal path.

\begin{figure}[h!]
  \centering
  \includegraphics[width=\linewidth]{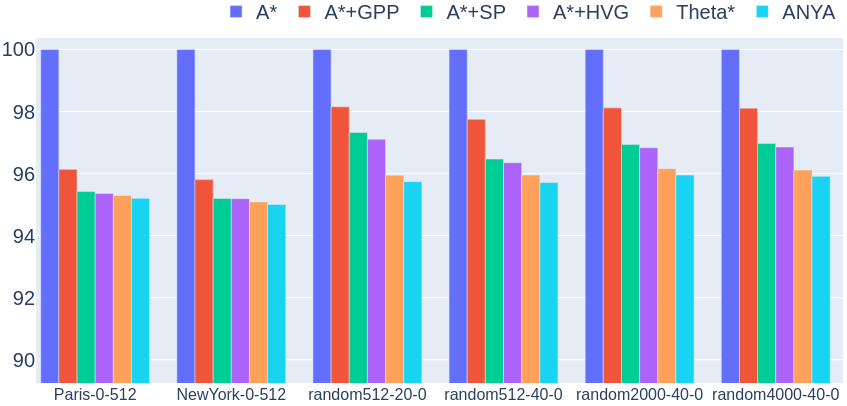}
  \caption{Path cost statistics with the grid-based path generated by A*. The random maps have names that contain the map size and the obstacle density in order. For example, random2000-40-0 is a randomly generated map of size 2000x2000 with 40\% obstacle density. The plot depicts the path cost as a percentage of the path cost of A* on the y-axis.}
  \label{fig:pathcost}
\end{figure}

\begin{figure}[h!]
  \centering
  \includegraphics[width=\linewidth]{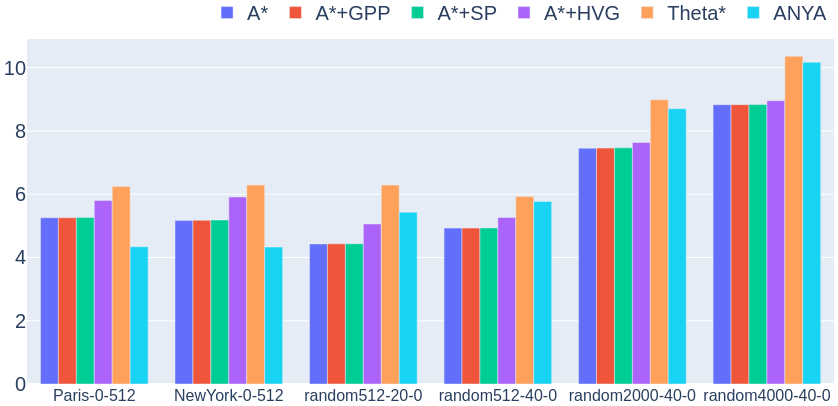}
  \caption{Runtime statistics with the grid-based path generated by A*. The random maps have names that contain the map size and the obstacle density in order. For example, random2000-40-0 is a randomly generated map of size 2000x2000 with 40\% obstacle density. The plot depicts the runtime (ms) in log-scale on the y-axis.}
  \label{fig:runtime}
\end{figure}

\noindent \textbf{Bounded Suboptimal Search + HVG: }Typically bounded suboptimal search methods like Weighted A* (wA*) are significantly faster than A* with the tradeoff on path cost. HVG can be used to post-process paths generated by wA* to provide homotopic Euclidean optimality guarantee while having a much smaller runtime compared to using A*. HVG provides a tradeoff opportunity by coupling HVG with weighted A* which worsens the path quality but the runtime improves significantly as the search component has at least a five-fold reduction in runtime.

\begin{figure}[h!]
    \centering
    \includegraphics[width=\linewidth]{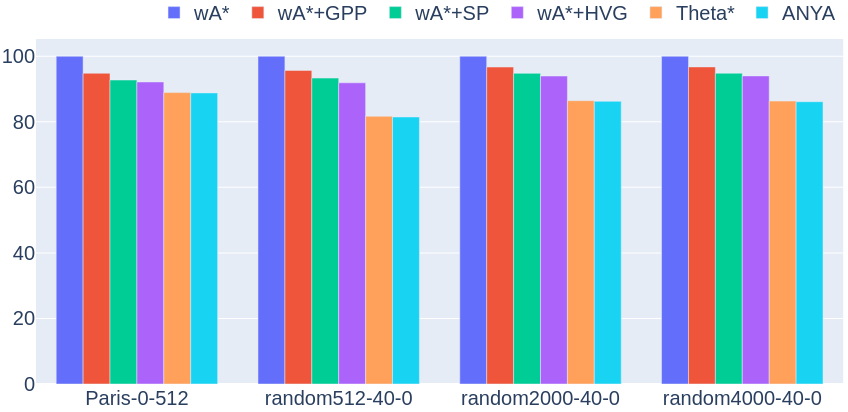}
    \caption{Path cost statistics with the grid-based path generated by wA* with w = 3. The plot depicts the path cost as a percentage of the path cost of wA* on the y-axis.}
    \label{fig:pathcostwt}
\end{figure}

\begin{figure}[h!]
    \centering
    \includegraphics[width=\linewidth]{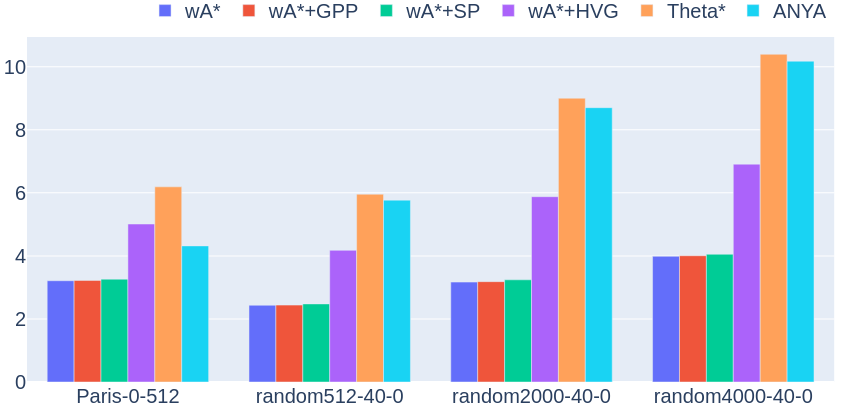}
    \caption{Runtime statistics with the grid-based path generated by wA* with w = 3. The plot depicts the runtime (ms) in log-scale on the y-axis.}
    \label{fig:runtimewt}
\end{figure}

In the case of the map `random4000-40-0', the cost of the path generated by A*+HVG is around 0.98\% worse than the path cost of ANYA with a 3x improvement in runtime over ANYA. Whereas the cost of the path generated by wA*+HVG is 9.1\% worse than that of ANYA with the runtime of wA*+HVG being 26x faster than ANYA. There is a substantial improvement in runtime when HVG is used along with weighted A*. The homotopic optimality guarantee still stands for wA*+HVG in the homotopy class of the grid-based path generated by wA*. 

\section{Conclusion \& Future Work}

In this work, we introduced a novel post-processing algorithm (HVG), which is, to the best of our knowledge, the first post-processing technique that returns a provably optimal path within the same homotopy class as the path returned by A* search run on a 2D grid. The algorithm is highly parallelizable and hence gives competitive runtimes. Other post-processing algorithms and any-angle algorithms do not lend themselves to be parallelized trivially. A$^*$ with HVG has shorter path lengths than A$^*$, A$^*$ with greedy post-processing and A$^*$ with String Pulling. In large maps and maps with high obstacle density, we demonstrated that A$^*$ with HVG consistently shows better runtimes than Theta$^*$ and ANYA. The runtime of A$^*$ with HVG is dominated by the time taken by A$^*$ for the initial grid search. However, HVG can also be used with weighted A$^*$. The runtime of the algorithm can be further improved by incorporating pruning techniques used in \cite{svg}. The parallelization of the vertex scanning and visibility graph construction was done using a naive allocation of one thread per node of the grid-based path. There is potential in improving multi-threading to get a sizeable reduction in runtime. 

\bibliography{references.bib}

\end{document}